\documentclass{article}
\usepackage{ijcai23}

\usepackage{times}
\usepackage{soul}
\usepackage{url}
\usepackage[hidelinks]{hyperref}
\usepackage[utf8]{inputenc}
\usepackage[small]{caption}
\usepackage{graphicx}
\usepackage{amssymb}
\usepackage{amsmath}
\usepackage{amsthm}
\usepackage{booktabs}
\usepackage{algorithmic}
\usepackage[switch]{lineno}

\usepackage{wrapfig}
\usepackage{enumerate}
\usepackage[ruled, vlined]{algorithm2e}%
\SetKwInOut{Parameter}{parameter}
\usepackage{float} 
\usepackage{subfigure,float}
\usepackage{mathrsfs}

\usepackage{bm}
\usepackage{dsfont}
\usepackage{tcolorbox,multicol}

\usepackage{threeparttable}
\usepackage{multirow}
\usepackage{cleveref}
\usepackage{caption}

\usepackage{paralist}
\usepackage{makecell}
\usepackage{pifont}
\usepackage{paralist}
\usepackage{diagbox}
\usepackage{float}
\usepackage{setspace}

\usepackage{subfigure,float}
\usepackage{wrapfig}
\usepackage{booktabs}
\usepackage{caption}

\usepackage{newfloat}
\usepackage{listings}

\usepackage{color}

\title{Controlling Neural Style Transfer with Deep Reinforcement Learning}

\author{
Chengming Feng${^1}$ \and 
Jing Hu${^1}$\and 
Xin Wang$^{2}$\thanks{Corresponding authors.}\and 
Shu Hu$^{3}$  \and 
Bin Zhu$^{4}$ \and 
Xi Wu${^{1}}^*$ \and \\
Hongtu Zhu$^{5}$ \And 
Siwei Lyu$^2$ \\
\affiliations
$^1$Chengdu University of Information Technology, China \\
$^2$University at Buffalo, SUNY, USA \\
$^3$Carnegie Mellon University, USA \\
$^4$Microsoft Research Asia, China \\
$^5$University of North Carolina at Chapel Hill, USA
\emails
xwang264@buffalo.edu,~
xi.wu@cuit.edu.cn
}

\begin{document}

\maketitle

\begin{abstract}
    Controlling the degree of stylization in the Neural Style Transfer (NST) is a little tricky since it usually needs hand-engineering on hyper-parameters.  
In this paper, we propose the first 
deep Reinforcement Learning (RL) based architecture that splits one-step style transfer into a step-wise process for the NST task. 
Our RL-based method tends to preserve more details and structures of the content image in 
early steps, and synthesize more style patterns in later steps. 
It is a user-easily-controlled style-transfer method. 
Additionally, as our RL-based model performs the stylization progressively, it is lightweight and has lower computational complexity than existing one-step Deep Learning (DL) based models. 
Experimental results demonstrate the effectiveness and robustness of our method. 
\end{abstract}

\section{Introduction}

Neural style transfer (NST)  refers to 
generation of a
pastiche image combining the semantic content of one image (the {\em content image}) and the visual style of another image (the {\em style image}) using a deep neural network. NST can be used to create 
stylized non-photorealistic rendering of digital images with enriched expressiveness and artistic flavors. 

Existing NST methods
usually generate a stylized image with a one-step approach: a neural network is trained to minimize a loss function of the visual similarity between the content image and the stylized image and the style similarity between the style image and the stylized image \cite{cheng2021style}, and the trained Deep Learning (DL) model is run once to create a stylized image.
This one-step approach has an apparent limitation: it is hard to determine a proper level of stylization to fit various flavors of different users since the ultimate metric of style transfer is very subjective. It is observed that generated stylized images by current NST methods tend to 
be under- or over-stylization~\cite{cheng2021style}. A remedy to 
under-stylization is to apply the DL model iteratively until a desired level of stylization is reached.
However, this solution 
may suffer from high computational cost due to intrinsic complexity of one-step DL models. Other existing methods, like~\cite{gatys2015neural} and \cite{huang2017arbitrary}, play a tradeoff between content and style by adjusting hyper-parameters. These methods are inefficient since there is no guarantee that a user can get the expected output via one-time adjusting.

\begin{figure*}[t]
    \includegraphics[width=1\textwidth]{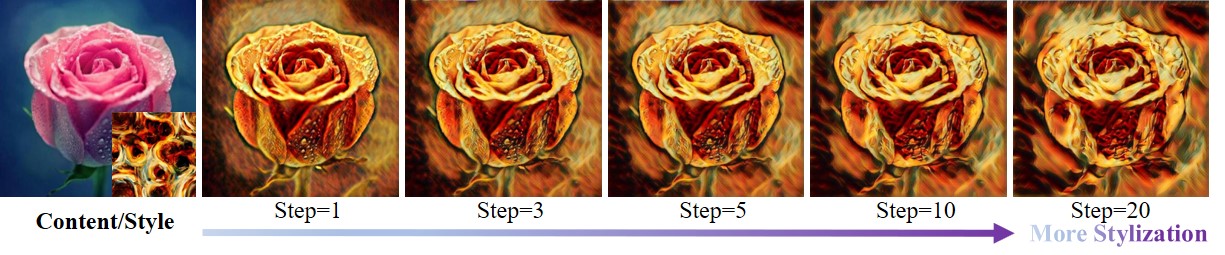}
    \caption{Illustration of our step-wise style transfer process. Content images are stylized smoothly stronger along with prediction steps. Our step-wise method can easily control the degree of stylization: the model tends to preserve more details and structures of the content image in early steps, and synthesize more style patterns in later steps. It is a user-easily-controlled style transfer method. 
    }
    \label{introdemo} 
\end{figure*}

\begin{figure*}[t]
    \includegraphics[width=1\textwidth]{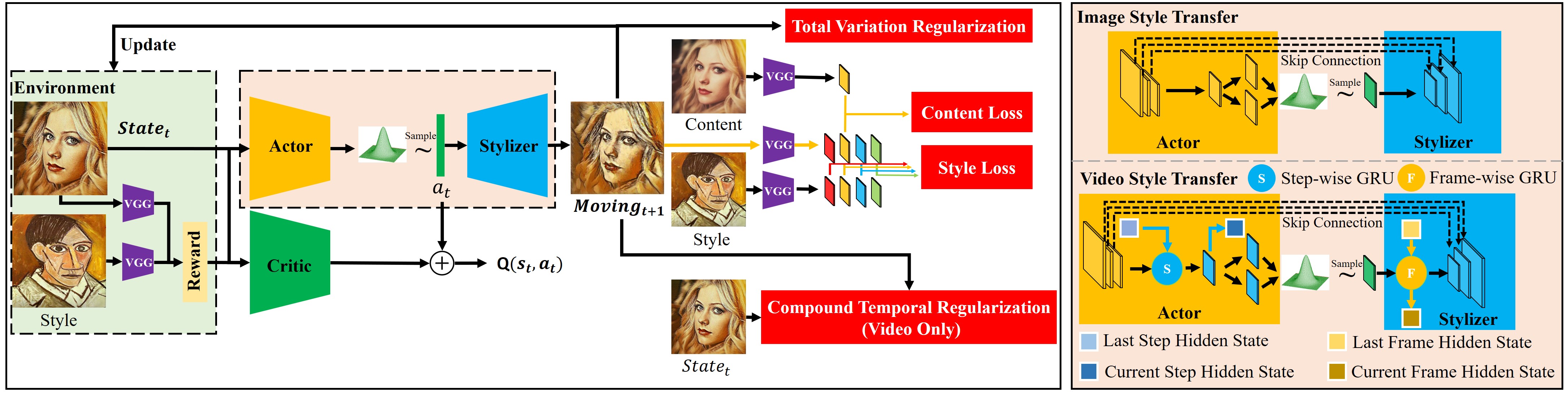}
    \caption{Our RL-NST framework. 
    \textbf{Left:}
    The state is initialized with the content image (or video frame). After the first iteration, we use only the moving image as the state. Latent-action ${\bf a}_t$ is sampled from a 2D Gaussian distribution and is concatenated with the critic's output. It is estimated by the policy $\pi_\phi$: ${\bf a_t} \sim \pi_\phi({\bf a}_t|{\bf s}_t)$. The predicted moving image is generated by stylizer $\eta_\psi$. Note that the VGG networks are pre-trained and fixed for the feature extraction during the training process. \textbf{Right:} The structure of the actor and stylizer for image and video NST, respectively.
    More details of the network structure can be found in Appendix.
    }
    \label{rlnst}
\end{figure*}

To address the aforementioned limitations of existing DL-based NST methods, we propose a novel framework in this paper, called \emph{RL-NST}, based on a reinforcement-learning (RL) framework to progressively perform style translation as shown in Fig. \ref{introdemo}. 
Given a content image, we consider the stylized content to be added progressively by the stylizer. 
More specifically, a stochastic actor in RL-NST first estimates a 2D Gaussian distribution to sample hidden actions, then uses the action to control the stylizer to generate an intermediate stylized image, which is in turn passed to the actor as the input for the next step.
Our model also includes a critic to evaluate the latent action. The whole structure is shown in Fig.~\ref{rlnst}. Furthermore, by using a CNN+RNN architecture~\cite{mirowski2016learning} for the actor and stylizer for both frame-wise and step-wise smoothing, our model can perform video NST tasks.
To the best of our knowledge, this is the first work that successfully leverages RL for the NST scenario. 

Our major contributions can be summarized as follows:
\textbf{1)} We propose the first reinforcement-learning-based NST method, RL-NST, that facilities step-wise style transfer. It provides more flexible control of the degree of stylization without any hyper-parameter adjustment during generation of stylized images. \textbf{2)}  Our RL-NST stylizes a content image progressively, with increased stylization along with more iterations (see Fig. \ref{introdemo}). It leads to a lightweight NST model compared with existing one-step DL-based methods, making it computationally more efficient. \textbf{3)} From our extensive experiments, our RL-NST demonstrates better effectiveness and robustness than existing state-of-the-art methods on both image and video NST tasks.





\section{Related Work}
\label{sec:bg}

\paragraph{Neural Style Transfer.}
Since the seminal work of Gatys et al. \cite{gatys2015neural} that uses a neural network to produce striking stylized art images, many methods have been proposed to improve the quality and/or running efficiency of NST algorithms. 
Existing NST methods can be divided roughly into two groups: image-optimization-based~\cite{gatys2015neural,li2017demystifying,risser2017stable} 
and model-optimization-based~\cite{johnson2016perceptual,ulyanov2016texture,chen2017stylebank,huang2017arbitrary,sheng2018avatar,park2019arbitrary,an2021artflow}. 

\paragraph{Reinforcement Learning.}
Reinforcement learning (RL) concerns how an agent takes actions in an environment to maximize its cumulative reward. Standard RL works well on tasks with finite and discrete action spaces. For real-world tasks with high dimensional continuous actions, such as robotic control~\cite{tassa2018deepmind,gu2017deep,xiang2022rmbench}, maximum entropy RL (MERL) and its variants, including Soft Q-learning~\cite{haarnoja2017reinforcement,zhao2019uncertainty} and SAC~\cite{haarnoja2018soft,hu2023attention}, have been proven to have a stable and powerful performance. But they still have limitations to handle 
high dimensional continuous state and action spaces in image-to-image (I2I) transformation and synthesis.
To address this problem, SAEC~\cite{luostochastic} extends the traditional MERL framework~\cite{haarnoja2018soft} with an additional executor. It shows promising performance on several I2I tasks.
However, its actions are 1D vectors, which do not preserve 2D spatial information of images. 


\section{Our RL-NST Framework}

We formulate NST as a decision-making problem. Instead of directly converting a content image to a stylized image in a single step, we propose to use a lightweight model to perform the translation progressively, with more stylization added to the stylized image as the translation progresses. 

Let $({\bf c}, {\bf e})$ be a pair of content and style images from image domain
$\mathcal{X}\in \mathbb{R}^d$, where $d$ is the dimension.
Our model consists of three components, as shown in Fig.~\ref{rlnst}: an actor $\pi_{\phi}$ parameterized by $\phi$, a stylizer $\eta_{\psi}$ with model parameters $\psi$, and a critic $Q_{\theta}$ parameterized by weights $\theta$. The actor generates a latent action according to a stochastic policy, so as to capture the style transfer control. 
The critic evaluates the generated action. The stylizer leverages the latent action and image details to perform style transfer on the content image and generate a stylized 
image, referred to as a \emph{moving image}, which updates the environment and is used subsequently as the content image in the next iteration. We use the pre-trained VGG \cite{simonyan2014very} as our feature extraction network because it is widely used such as in \cite{chen2021artistic,lin2021drafting}. 
Actor $\pi_{\phi}$ and stylizer $\eta_{\psi}$ are supervised jointly with content loss,   style loss, total variation regularization, and compound temporal regularization (for video only) on the current content image (i.e., the moving image). Furthermore, the actor $\pi_{\phi}$ and critic $Q_{\theta}$ form an actor-critic model. 

\subsection{RL-NST Settings}
\label{sec:i2i-mdp}
 In forming our model, we define the infinite-horizon Markov decision process (MDP) as  tuple $({\cal S},{\cal A}, {\cal P},{r}, {\gamma})$, where ${\cal S}$ is a set of states, ${\cal A}$ is the action space, and  ${\cal P}: {\cal S}  \times {\cal S} \times {\cal A} \rightarrow [0, \infty)$ represents the state transition probability of the next state ${\bf s}_{t+1}$ given ${\bf s}_t \in {\cal S}$ at time $t$ and action ${\bf a}\in {\cal A}$, ${r}: {\cal S} \times {\cal A} \rightarrow \mathbb{R}$ is the reward emitted from each transition, and $\gamma \in [0,1]$ is the reward discount factor. Specifically,
(1) \textbf{State}. 
State ${\bf s}_t$ is the moving image, initialized by the content image. (2) \textbf{Action}. To extract high-level abstraction ${\bf a}_t$ of ${\bf s}_t$ from the actor, a stochastic latent action is modeled as ${\bf a}_t\sim \pi_\phi({\bf a}_t|\textbf{s}_t)$. In practice, it can be obtained by using a reparameterization trick~\cite{kingma2013auto} ${\bf a}_t=f_\phi(\epsilon_t, {\bf s}_t)$, where $\epsilon_t$ is an input noise vector sampled from a 2D Gaussian distribution. The moving image at time $t$, i.e., state image ${\bf s}_{t+1}$, is created by the stylizer based on ${\bf a}_t$ and current state image ${\bf s}_t$. (3) \textbf{Reward}. It is from the environment $\cal E$, obtained by measuring the difference between current state ${\bf s}_t$ and the style image. The higher the difference is, the smaller the reward is.


\subsection{Network Architecture }

The \textbf{Actor} is a neural network model that consists of three convolutional layers and a residual layer. After each convolutional layer, there is an instance norm layer and a ReLU layer. In the residual layer, we use the residual block designed by He et al.~\cite{he2016deep}.
The actor estimates a 2D Gaussian distribution for sampling our latent actions, which is forwarded to the stylizer to generate the moving image.
The dimension of our latent action is 64$\times$64, which is a 2D sample and able to preserve more spatial structure information of images. 
Notably, the actor in our method is to learn the latent actions, which are more compact than the image presentation. 
The learned actions are controlled by our framework to guide the stylizer to generate stylized images instead of the reconstruction.

The \textbf{Stylizer} has three up-sampling layers correspondingly.  
More importantly, by using 2D Gaussian sampling, our actor-stylizer structure is a fully convolutional network (FCN), which can process images of any input size, instead of only accepting test images of the same size as training images. We also use three skip connections between the actor and the stylizer to stabilize the training process. 

To handle video NST tasks, we expand the actor-stylizer to include RNN layers by following the work~\cite{mirowski2016learning}. In particular, the ConvGRU layer~\cite{shi2015convolutional} is used for the FCN structure (see the right bottom of Fig.~\ref{rlnst}). We add Step-wise GRU~\cite{mirowski2016learning} to actor and Frame-wise GRU~\cite{7558228} to stylizer. 
Specifically, for each frame, the hidden state of the Step-wise GRU at each step comes from the output of the Step-wise GRU at the previous step. The role of Step-wise GRU is to make the model maintain better content consistency. Furthermore, the hidden state of the Frame-wise GRU at each step is derived from the output of the Frame-wise GRU at the same step in the previous frame. Frame-wise GRU can make the model maintain better inter-frame consistency.

The \textbf{Critic} consists of seven convolutional layers and one fully-connected layer at the end. 
Since 
using standard zero-padded convolutions in style transfer leads to serious artifacts on the boundary of a generated image~\cite{ulyanov2016instance}, we use reflection padding instead of zero padding for all the networks.

\subsection{Model Training}
\label{sec:training}
Our RL-NST contains two learning procedures, namely style learning and step-wise learning.

\subsubsection{Style Learning}  
To make the moving image not deviate from the content image, 
actor $\pi_\phi$ (encoder) and stylizer $\eta_\psi$ (decoder) are trained together to preserve the perceptual and semantic similarity with the content image.  
More specifically, the actor and the stylizer form a conditional generative process that translates state ${\bf s}_t$ to output moving image ${\bf m}_t$ via the mapping ${\bf m}_t=\eta_\psi(\pi_\phi({\bf s}_t))$ at time $t$. Note that ${\bf s}_t$ is initialized to content image ${\bf c}$ and ${\bf s}_{t+1}$ 
is equivalently ${\bf m}_t$.
Inspired by  \cite{johnson2016perceptual}, we apply the content loss $\mathcal{L}^{CO}$, style loss $\mathcal{L}^{ST}$, and total variation regularization $\mathcal{L}^{TV}$ to optimize the model parameters of $\pi_\phi$ and $\eta_\psi$ in the image setting. These losses can better measure perceptual and semantic differences between the moving image and content image ${\bf c}$. For the video setting, we add an additional loss named compound temporal regularization $\mathcal{L}^{CT}$, which can force the model to generate temporal consistent results under the compound transformation. More details about these losses are as follows.

\paragraph{Content Loss $\mathcal{L}^{CO}$.}
Following \cite{johnson2016perceptual}, we use a pre-trained neural network $F$ to extract the high-level feature representatives of ${\bf m}_t$ and ${\bf c}$. The reason for using this $F$ is to encourage moving image ${\bf m}_t$ to be perceptually similar to content image ${\bf c}$ but does not force them to match exactly. Denote $F^j(\cdot)$ as the activations of the $j$-th layer of $F$. Suppose $j$-th layer is a convolutional layer, then the output of $F^j(\cdot)$ will be a feature map with size $C^j\times H^j\times W^j$, where $C^j$, $H^j$, and $W^j$ represent the number of channels, height, and width in the feature map of layer $j$, respectively. We apply the Euclidean distance, which is squared and normalized to design the content loss as follows,
\begin{equation*}
    \mathcal{L}^{CO} ({\bf m}_t, {\bf c}) = \frac{1}{C^jH^jW^j}\|F^j({\bf m}_t)-F^j({\bf c})\|_2^2.
\end{equation*}

\paragraph{Style Loss $\mathcal{L}^{ST}$.}
To penalize ${\bf m}_t$ when it deviates in content from ${\bf c}$ and in style from ${\bf e}$, following \cite{gatys2015texture}, we define a Gram matrix $G^j({\bf x})=\frac{\tilde{F}^j({\bf x})(\tilde{F}^j({\bf x}))^\top}{C^jH^jW^j}\in \mathbb{R}^{C^j\times C^j}$, where $\tilde{F}^j(\cdot)$ is obtained by reshaping $F^j(\cdot)$ into the shape $C^j\times H^jW^j$. The style loss can be defined as a squared Frobenius norm of the difference between the Gram matrices of ${\bf m}_t$ and ${\bf e}$. To preserve the spatial structure of images, we use a set of layers, $J$, instead of a single layer $j$. Thus, we define the style loss to be the sum of losses for each layer $j\in J$ ($J=4$ in our experiments):
\begin{equation*}
    \mathcal{L}^{ST} ({\bf m}_t, {\bf e}) = \sum_{j=1}^J\|G^j({\bf m}_t)-G^j({\bf e})\|_F^2.
\end{equation*}

\paragraph{Total Variation Regularization $\mathcal{L}^{TV}$.}
To ensure spatial smoothness in moving image ${\bf m}$, we use a total variation regularizer $\mathcal{L}^{TV}({\bf m}_t)$, which has been widely used in existing works \cite{mahendran2015understanding,johnson2016perceptual}. 

\paragraph{Compound Temporal Regularization $\mathcal{L}^{CT}$.}
Inspired by \cite{wang2020consistent}, we add a compound temporal regularization for video style transfer. Specifically, we first generate motions $M(\cdot)$ and then synthesize adjacent frames. With this approach, we do not need to estimate optical flow in the training process and we can guarantee the optical flows are absolutely accurate. Given noise $\triangle$, to maintain temporal consistency, we can minimize the following loss
\begin{equation*}
    \mathcal{L}^{CT}=\|\eta_\psi(\pi_\phi(M({\bf s}_t)+\triangle))-M({\bf m}_t)\|_1.
\end{equation*}

Summing up all the components, the final style learning loss is 
\begin{equation}
    \mathcal{L} = \underbrace{\overbrace{ \mathcal{L}^{CO}+\lambda\mathcal{L}^{ST}+\beta\mathcal{L}^{TV}}^{\text{for image}}+\zeta\mathcal{L}^{CT}}_{\text{for video}},
    \label{eqlst}
\end{equation}
where $\lambda$, $\beta$, and $\zeta$ are hyper-parameters to control the sensitivity of each term. For image style transfer, we use the first three terms. For video style transfer, we use all terms. Then we can update $\phi$ and $\psi$ from the actor and stylizer by using the gradient descent method with a predefined learning rate $\eta$ with the following steps:
\begin{equation}
    \phi \leftarrow \phi-\eta\nabla_{\phi}\mathcal{L}, \ \ \ \psi \leftarrow \psi-\eta\nabla_{\psi}\mathcal{L}.
\label{eq:dl}
\end{equation}

\subsubsection{Step-wise Learning} 
Our step-wise learning is based on the MERL framework~\cite{haarnoja2018soft}, where  rewards and soft Q values are used to iteratively guide the stochastic policy improvement. Moreover, we focus on latent action ${\bf a}$, 
and use it to estimate a soft state-action value to encourage high-level policies. Specifically, we concatenate ${\bf a}_t$ to the downsampled vector of the critic and output soft Q function $Q_{\theta}({\bf s}_t, {\bf a}_t)$, which is an estimation of the state value at time $t$. Because the critic is used to evaluate the actor, rewards $r_t$ and the soft Q values are used to iteratively guide the stochastic policy improvement by minimizing the soft Bellman residual:
\begin{equation*}
    \begin{aligned}
    J_Q(\theta) = 
    \mathbb{E}_{({\bf s}_t,{\bf a}_t)\sim  \mathcal{D}} \big[\frac{1}{2} \Big(Q_{\theta}({\bf s}_t, {\bf a}_t) - \\ \big(r_t + \gamma \mathbb{E}_{{\bf s}_{t+1}}\left[V_{\bar{\theta}}({\bf s}_{t+1})\right]\big)\Big)^2 \big],
    \end{aligned}
\end{equation*}

where $\mathcal{D}$ is a replay pool and $V_{\bar{\theta}}({\bf s}_{t}) = \mathbb{E}_{{\bf a}_{t}\sim \pi_\phi}[Q_{\bar{\theta}}({\bf s}_{t}, {\bf a}_{t}) - \alpha \log \pi_\phi ({\bf a}_{t}|{\bf s}_{t})]$. 
Note that we utilize the negative value of $\mathcal{L}^{ST}({\bf s}_t, {\bf e})$ for $r_t$ in practice.

\begin{algorithm}[t]
    \caption{RL-NST}\label{Alg0}
    \SetAlgoLined

    \KwIn{${\bf c}$, ${\bf e}$, and replay pool $\mathcal{D}$}

    \textbf{Init:} $\phi$, $\psi$, $\theta$, $\bar{\theta}$, $\mathcal{D}\leftarrow \emptyset$, $\eta$, $\eta_Q$, $\eta_\phi$, and environment $\cal E$
    
    \For{each iteration}{
    \For{each environment step}{
    ${\bf a}_{t} \sim \pi_\phi({\bf a}_{t} | {\bf s}_{t})$
    
    ${\bf s}_{t+1}, {r_t} \sim \mathcal{P}({\bf s}_{t+1}|{\bf s}_{t},  {\bf a}_{t})$
    
    $\mathcal{D} \leftarrow \mathcal{D} \cup \{({\bf s}_{t}, {\bf a}_{t}, r_t, s_{t+1})\}$
    }
    
    \For{each gradient step}{
    Sample from $\mathcal{D}$
    
    Update $\theta$, $\phi$, $\psi$ by using Eq.(\ref{eq:dl}), (\ref{eq:update_theta}), and (\ref{eq:update_phi})
    }
    }
\end{algorithm}

The critic network $Q_{\bar{\theta}}$ is used to stabilize the training, whose parameters $\bar{\theta}$ are obtained by an exponential moving average of parameters of the critic network \cite{lillicrap2015continuous}: $\bar{\theta} \rightarrow \tau \theta + (1-\tau)\bar{\theta}$, with hyperparameter $\tau\in [0,1]$. To optimize $J_Q(\theta)$,  we use the gradient descent with respect to parameters $\theta$ as follows,
\begin{equation*}
    \begin{aligned}
    \theta \leftarrow& \theta - \eta_Q \triangledown_{\theta} Q_{\theta}({\bf s}_t, {\bf a}_t)\Big(Q_{\theta}({\bf s}_t, {\bf a}_t) \\
    -& r_t - \gamma \left[Q_{\bar{\theta}}({\bf s}_{t+1}, {\bf a}_{t+1}) - \alpha \log \pi_{\phi} ({\bf a}_{t+1}|{\bf s}_{t+1})\right]\Big),
    \end{aligned}
\label{eq:update_theta}
\end{equation*}
where $\eta_Q$ is a learning rate. Since the critic works on the actor, it will affect the actor's decisions. Therefore, the following objective can be applied to minimize the KL divergence between the policy and a Boltzmann distribution induced by the Q-function, 

\begin{equation*}
    \begin{aligned}
    J_{\pi} (\phi) =& \mathbb{E}_{ {\bf s}_t\sim  \mathcal{D}}\big[ \mathbb{E}_{{\bf a}_t\sim  \pi_{\phi}} \left[\alpha \log (\pi_{\phi}({\bf a}_t| {\bf s}_t))-Q_\theta({\bf s}_t,{\bf a}_t)\right]\big]\\
    =& \mathbb{E}_{ {\bf s}_t\sim  \mathcal{D}, {\bf \epsilon}_t\sim  \mathcal{N} (\bm{\mu},\bm{\Sigma}) } \big[\alpha \log (\pi_{\phi}( f_\phi({\bf \epsilon}_t,{\bf s}_t)|{\bf s}_t))\\
    -& Q_\theta({\bf s}_t,f_\psi({\bf \epsilon}_t,{\bf s}_t))\big].
    \end{aligned}
\end{equation*}

\begin{figure*}[ht]
    \includegraphics[width=1\textwidth]{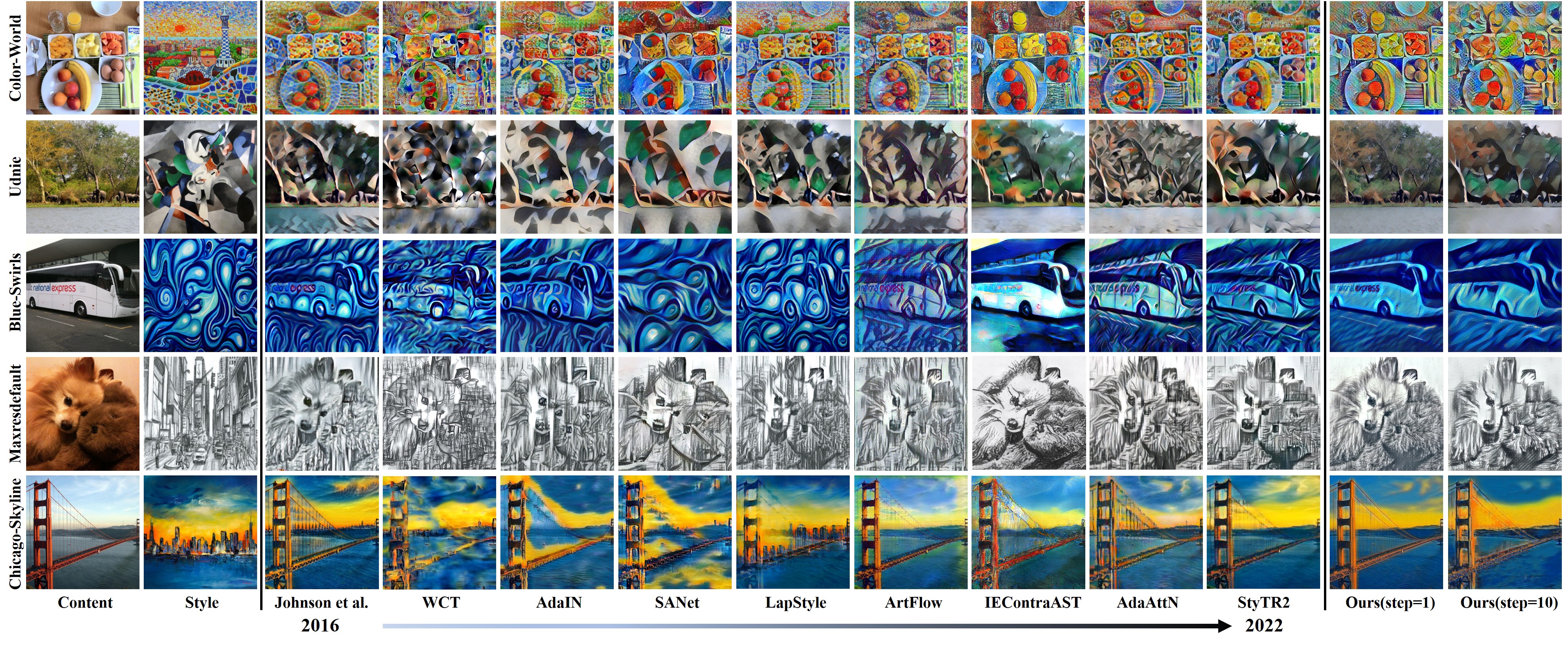}
    \caption{Qualitative comparison. The first two columns show the content and style images, respectively. The rest of the columns show the stylization results generated with different style transfer methods, with the last two columns of our step-wise results at step 1 and 10, respectively.}
    \label{visualcompare}
\end{figure*}

\begin{table*}[t]
\centering
\scalebox{0.715}{
\begin{tabular}{c|ccccccccc|cc}
\hline
Methods      & Johnson et al.   & AdaIN     & WCT       & SANet &LapStyle    & ArtFlow   & IEContraAST & AdaAttN   & StyTR2    & Ours(step=1) & Ours(step=10) \\ \hline
Content loss & 1.597     & 2.222     & 2.322     & 1.941 &2.292    & 1.538     & 1.668       & 1.447     & 1.510     & \textbf{0.868}        & \textbf{1.387}         \\ \hline
Style loss   & 1.985e-05 & 1.269e-05 & 1.626e-05 & 7.062e-06 &2.117e-05 & 1.486e-05 & 8.863e-06   & 1.033e-05 & 9.178e-06 & \textbf{3.353e-06}    & \textbf{1.594e-06}     \\ \hline

Time (s)         & \makecell{0.014 \\(3.5$\times$)}  & \makecell{0.140 \\(35$\times$)} &\makecell{0.690\\($172.5\times$)}   & \makecell{0.010\\ (2.5$\times$)} & \makecell{0.047\\ (11.75$\times$)}    & \makecell{0.127\\ (31.75$\times$)}  & \makecell{0.019\\ (4.75$\times$)} & \makecell{0.025\\(6.25$\times$)}  & \makecell{0.058\\(14.5$\times$)}   & \textbf{0.004} & 0.089                 \\\hline
\#Params (M)     & \makecell{1.68 \\(9.33$\times$)}  & \makecell{7.01 \\(38.94$\times$)} &\makecell{34.24 \\(190.22$\times$)}  & \makecell{20.91 \\ (116.17$\times$)}  & \makecell{7.79 \\ (43.28$\times$)}  & \makecell{6.46 \\(35.89$\times$)}   & \makecell{21.12 \\(117.33$\times$)} & \makecell{13.63 \\(75.72$\times$)}  &\makecell{35.39 \\(196.61$\times$)} &\textbf{0.18} & \textbf{0.18}        \\\hline

\end{tabular}}
\captionof{table}{Quantitative comparison of our RL-NST with the baseline methods on the MS-COCO dataset. The speed is obtained with a Pascal Tesla P100 GPU. ($\cdot\times$) represents the ratio between current baseline and our method (step=1) under the same metric. The best results
are shown in bold. 
}
\label{avgnumbercompare}
\end{table*}

The last equation holds because ${\bf a}_t$ can be evaluated by $f_\phi({\bf \epsilon}_t,{\bf s}_t)$, as we discussed before.Note that hyperparameter $\alpha$ can be automatically adjusted by using the method proposed in~\cite{haarnoja2018soft}. Similarly, we apply the gradient descent method with a learning rate $\eta_\phi$ to optimize parameters as follows,
\begin{equation}
    \begin{aligned}
    \phi \leftarrow& \phi -\eta_\phi\Big(\triangledown_\phi \alpha \log(\pi_\phi({\bf a}_t|{\bf s}_t)) + \big(\triangledown_{{\bf a}_t}\alpha \log(\pi_{\phi}({\bf a}_t|{\bf s}_t))\! \\
    -&\!\triangledown_{{\bf a}_t}Q_{\theta}({\bf s}_t,{\bf a}_t)\big) \triangledown_\phi f_\phi (\epsilon_t,{\bf s}_t)\Big).
    \end{aligned}
\label{eq:update_phi}
\end{equation}

The pseudo-code of optimizing RL-NST is described in Algorithm \ref{Alg0}. All parameters are optimized based on the samples from replay pool $\cal D$.

\section{Experiments}

We have conducted a series of experiments to evaluate the effectiveness of RL-NST in realizing step-wise style transfer on both image and video NST tasks. Our \textbf{code}, a user study, and additional results with more detailed information can be found in the supplementary materials.

\subsection{Experimental Settings}

\paragraph{Datasets.}
(1) For image style transfer, we select style images from WikiArt~\cite{phillips2011wiki} and use MS-COCO \cite{lin2014microsoft} as content images in which the training set includes 80K images and the test set includes 40K images.
All training images are resized to 256$\times$256. In the inference stage, our method is applicable for content images and style images of any size.
(2) For video style transfer, we randomly collect 16 videos of different scenes from pexels\cite{pexels2022}. Then these videos are extracted into video frames and we obtain more than 2.5K frames. We regard these frames as the content images of training set. Note that the style images in the training set are also selected from WikiArt~\cite{phillips2011wiki}. In addition, following \cite{wang2020consistent}, we use the training set of MPI Sintel dataset~\cite{butler2012naturalistic} as the test set, which contains 23 sequences with a total of 1K frames.
Similarly, all training frames are resized to 256$\times$256, we use the original frame size in testing.

\paragraph{Baselines and Evaluation Metrics.}
(1) For image style transfer, we choose the following eight classical and latest state-of-the-art style transfer methods as our baselines: 
Johnson et al.~\cite{johnson2016perceptual},
WCT~\cite{li2017universal},
AdaIN~\cite{huang2017arbitrary}, 
SANet~\cite{park2019arbitrary}, 
LapStyle~\cite{lin2021drafting},
ArtFlow~\cite{an2021artflow},
IEContraAST~\cite{chen2021artistic}, AdaAttN~\cite{liu2021adaattn},
and StyTR2~\cite{deng2021stytr}.
Following StyTR2 \cite{deng2021stytr}, we evaluate all the algorithms in terms of stylization effect, computing time, content loss, and style loss. 
(2) For video style transfer, we compare our method with the following four popular methods: Linear~\cite{li2019learning}, MCCNet~\cite{deng2020arbitrary}, ReReVST~\cite{wang2020consistent}, and AdaAttN~\cite{liu2021adaattn}.
Following \cite{liu2021adaattn}, we use temporal loss as the evaluation metric to compare the stability of stylized results. All these methods are performed using their public codes with the default settings.

\paragraph{Implementation Details.}
In the experiment, we set $\lambda = 1e5$, $\beta = 1e-7$, $\zeta=1e2$ in Eq. (\ref{eqlst}), and $\eta=1e-4$ in Eq. (\ref{eq:dl}). These settings yield nearly the
best performance in our experiments. Following \cite{wang2020consistent}, in $\mathcal{L}^{CT}$, $M(\cdot)$ is implemented by warping with a random optical flow. Specifically, for a frame of size $H\times W$, we first generate a Gaussian map (wavy twists) $M_{wt}$ of shape $H/100\times W/100 \times 2$, mean 0, and standard deviation 0.001. Second, $M_{wt}$ is resized to $H\times W$ and blurred by a Gaussian filter of kernel size 100. Finally, we add two random values (translation motion) $M_{tm}$ of range [-10,10] to $M_{wt}$, and obtain $M$. In addition, random noise $\triangle \sim \mathcal{N} (0,\sigma^2I)$, where $\sigma \sim \mathcal{U}(0.001, 0.002)$.

\begin{table}[htb]
    \centering
            \scalebox{0.78}{
            \begin{tabular}{c|c|c|cc}
            \toprule
                                  Style                &Step                   &Method        &  Content Loss    & Style Loss  \\ \midrule
             \multirow{4}{*}{Maxresdefault}  &  \multirow{2}{*}{1}    & Actor-Stylizer (AS)  & 0.787                      & 5.686e-06   \\
                                                                    &    & \textbf{Ours}& \textbf{0.557}             & \textbf{1.883e-06}  \\\cline{2-5} 
                                                     & \multirow{2}{*}{10}     & Ours~w/o RL  & 2.093                      & 3.433e-05 \\
                                                                          &    & \textbf{Ours } & \textbf{0.945}             & \textbf{1.093e-06}  \\ \hline
             \multirow{4}{*}{Blue Swirls}            & \multirow{2}{*}{1}      & Actor-Stylizer (AS) & 2.265                      & 3.275e-05   \\
                                                                            &   & \textbf{Ours} & \textbf{1.016}             & \textbf{5.280e-06} \\\cline{2-5} 
            
                     & \multirow{2}{*}{10}    & Ours~w/o RL & 3.374                      & 7.747e-05 \\
                      &   & \textbf{Ours }  & \textbf{1.654}             & \textbf{2.178e-06}  \\\hline
            \end{tabular}}
            \captionof{table}{ Content loss, and style loss of several variants of our proposed method. 
            }
            \label{tab_abl}
\end{table}

\begin{figure}[htb]
    \centering
    \includegraphics[scale=0.23]{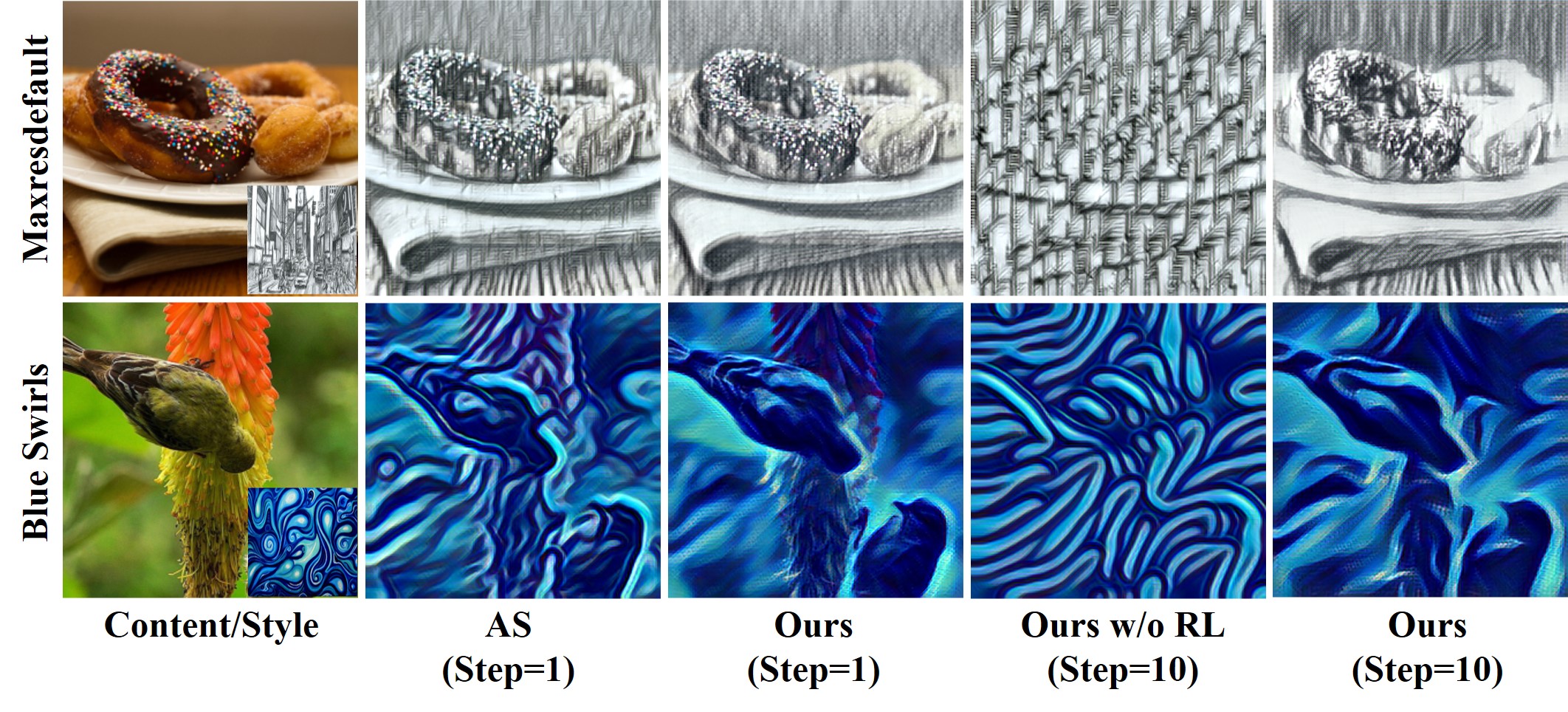}
    \caption{Comparison of Ours with the Actor-Stylizer (AS) model at step 1 and Ours without the RL model at step 10.}
    \label{compare_vae}
\end{figure}

\begin{figure}[t]
    \centering
    \includegraphics[scale=0.24]{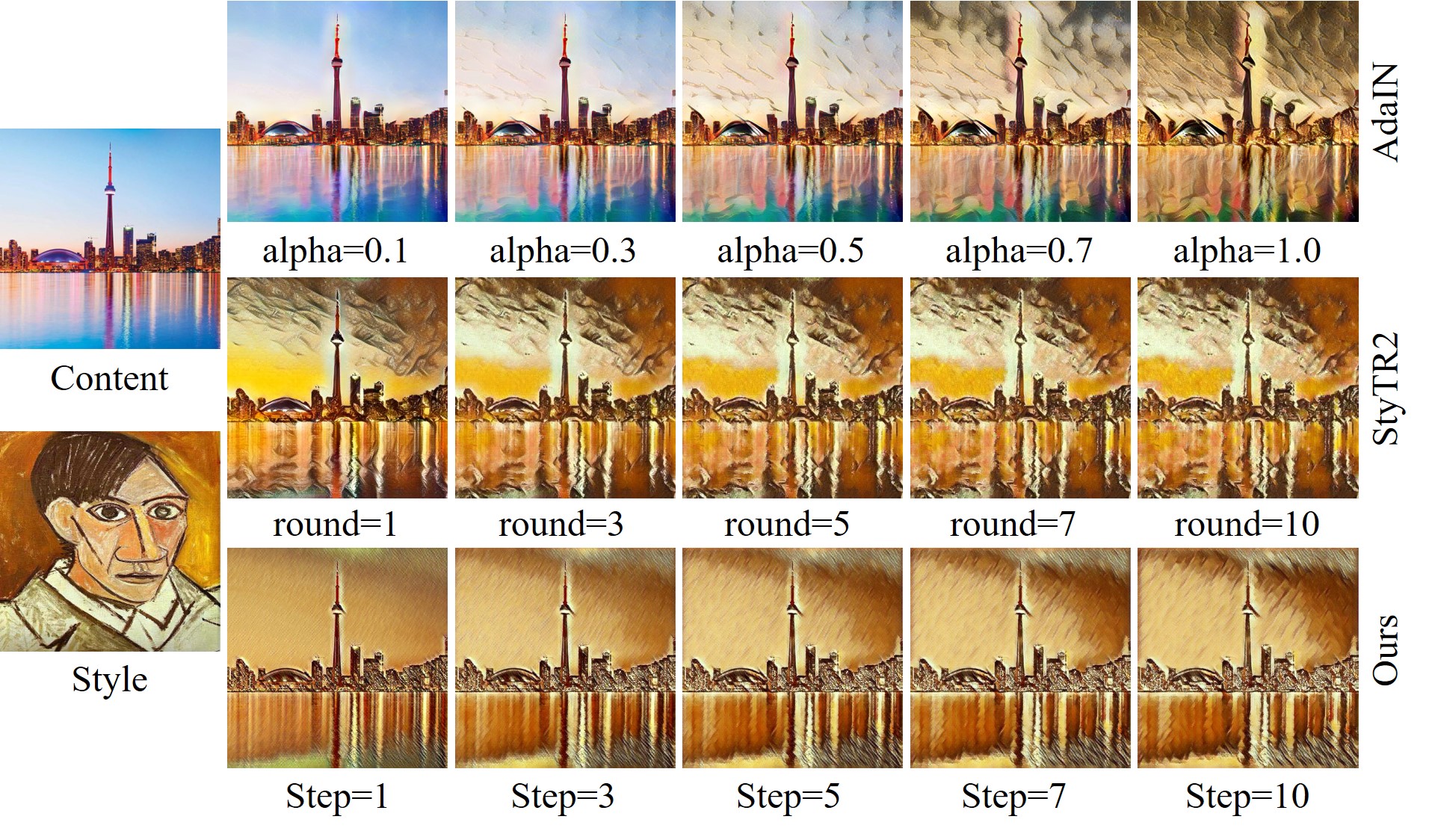}
    \caption{Comparison of Ours with AdaIN and StyTR2 in various hyperparameter settings.}
    \label{compare_adain_stytr2}
\end{figure}

\subsection{Evaluations on Image RL-NST}

\paragraph{Qualitative Comparison.}
Fig.~\ref{visualcompare} shows some stylized results of our RL-NST and the baseline methods. 
For content images with fine structures such as the forest image (Udnie style), all the baseline methods proposed to address content leakage, including ArtFlow, produce messy stylized images with a complete loss of content structure.
Moreover, SANet has repeated texture patterns for all the cases, and most of its results are hard to generate sharp edges.

In contrast, our method can produce stable and diversified stylized results with good content structures. This may be attributed to our step-wise solution. More specifically, the content image is stylized progressively and hence smoothed stylization results are obtained. More importantly, as we mentioned before, despite that stylization is quite subjective, our step-wise method provides flexible control of different degrees of stylization to fit the need of different users.

\paragraph{Quantitative Results.}
To be consistent with all compared methods shown in Fig.~\ref{visualcompare}, we compare our method with all baselines without caring which type (single or multiple styles) they are.   
The quantitative results are shown in Table~\ref{avgnumbercompare}. Our RL-NST (step=1) achieves better performance than the baseline methods in all evaluation metrics. Our method still has low content and style losses even if the step is equivalent to 10, which means our method is friendly to the user for choosing the results from specific steps accordingly.    
In addition, it is clear that our model has much fewer parameters and a faster speed. For example, the time cost and the parameter size of our method are $2/7$ and $1/9$ of Johnson et al., and $4/47$ and $1/43$ of LapStyle, respectively.

\begin{figure}[tb]
    \centering
    \includegraphics[scale=0.135]{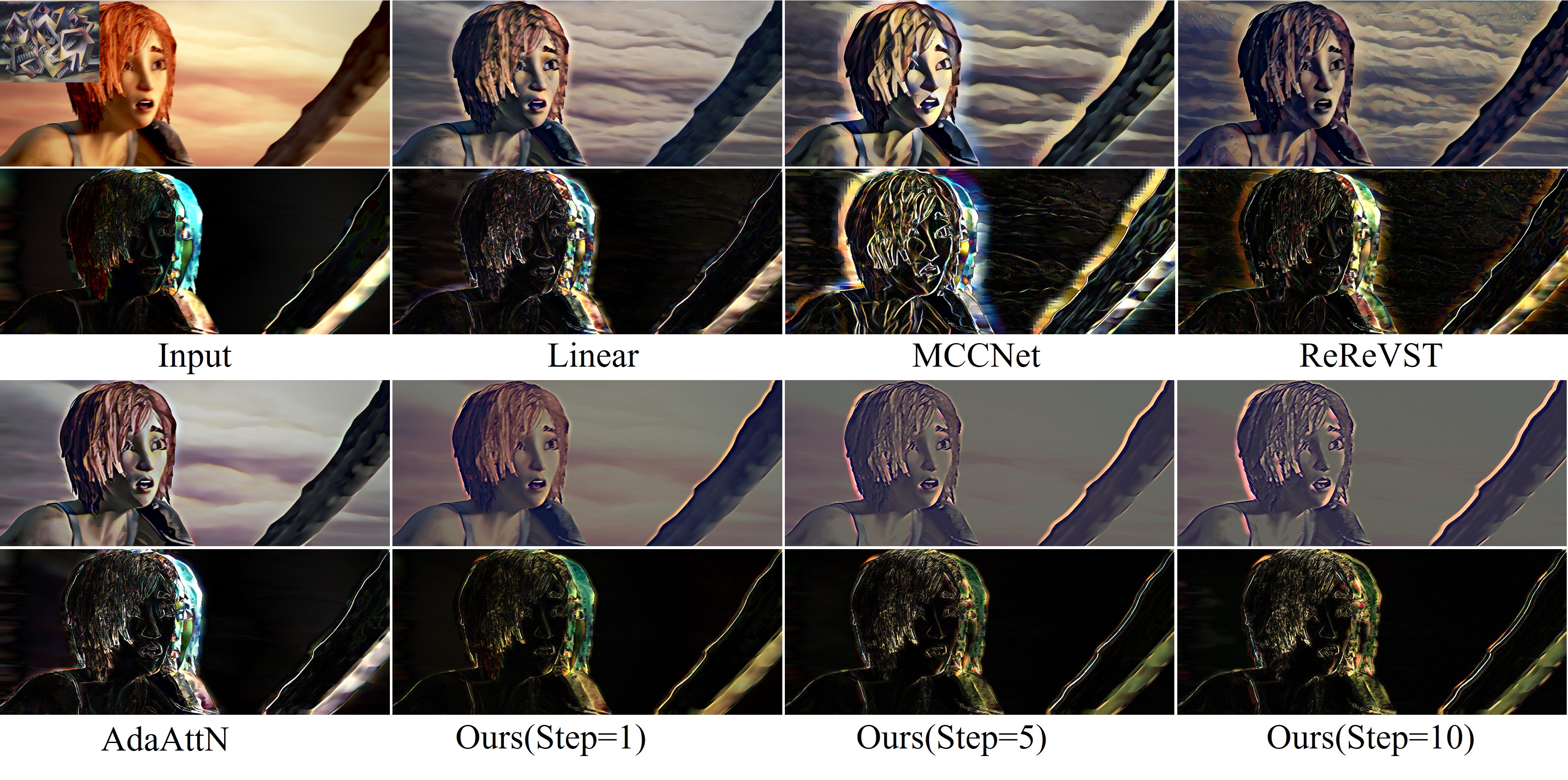}
    \caption{Comparison of video style transfer between our method and the compared methods. For each method, the top portion shows the video frame stylized results. The bottom portion shows the heatmap of the differences between two adjacent video frames.
    }
    \label{fig:compare_video}
\end{figure}

\begin{table*}[t]
\centering
\small{
\begin{tabular}{c|cccccccc|c}
\hline
\diagbox{Methods}{Styles}& La\_muse & Sketch  & En\_campo\_gris & Brushstrokes & Picasso & Trial   & Asheville & Contrast &Average \\ \hline
LinearStyleTransfer   & 2.602           & 1.792          & 1.795         & 2.321          & 2.947         & 1.451           & 5.043            & 4.524 &2.809  \\
ReReVST               & 1.450           & 8.155          & 7.050          & 7.026          & 10.772         & 7.888          & 19.493            & 12.886 &9.340  \\
MCCNet                & 4.493           & 2.050          & 2.759          & 2.591         & 2.854          & 2.486          & 6.750            & 4.820 &3.600  \\
AdaAttN               & 3.442           & 1.976          & 2.660          & 2.561          & 2.941          & 1.698          & 5.775            & 3.587 &3.080  \\ \hline
Ours(Step=1)          & \textbf{0.885}  & \textbf{1.196} & \textbf{0.453} & \textbf{0.883} & \textbf{1.447} & \textbf{0.527} & \textbf{1.735}   & \textbf{1.045} &\textbf{1.021}  \\
Ours(Step=5)          & \textbf{1.436}  & \textbf{1.509} & \textbf{0.855} & \textbf{1.499} & \textbf{1.980} & \textbf{0.704} & \textbf{2.327}   & \textbf{1.550} &\textbf{1.483} \\
Ours(Step=10)         & 1.867           & \textbf{1.695} & \textbf{1.141} & \textbf{1.807}  & \textbf{2.394} & \textbf{0.852} & \textbf{2.854}   & \textbf{1.842} &\textbf{1.807}  \\ \hline
\end{tabular}}
\captionof{table}{Comparison of the average temporal losses ($\times 10^{-2}$) from 23 different sequences of our method with other baseline methods on different styles. The last column shows the average scores among all styles in each method. 
}
\label{tab:compare_video_methods}
\end{table*}

\begin{figure*}[htb]
    \centering
    \includegraphics[scale=0.23]{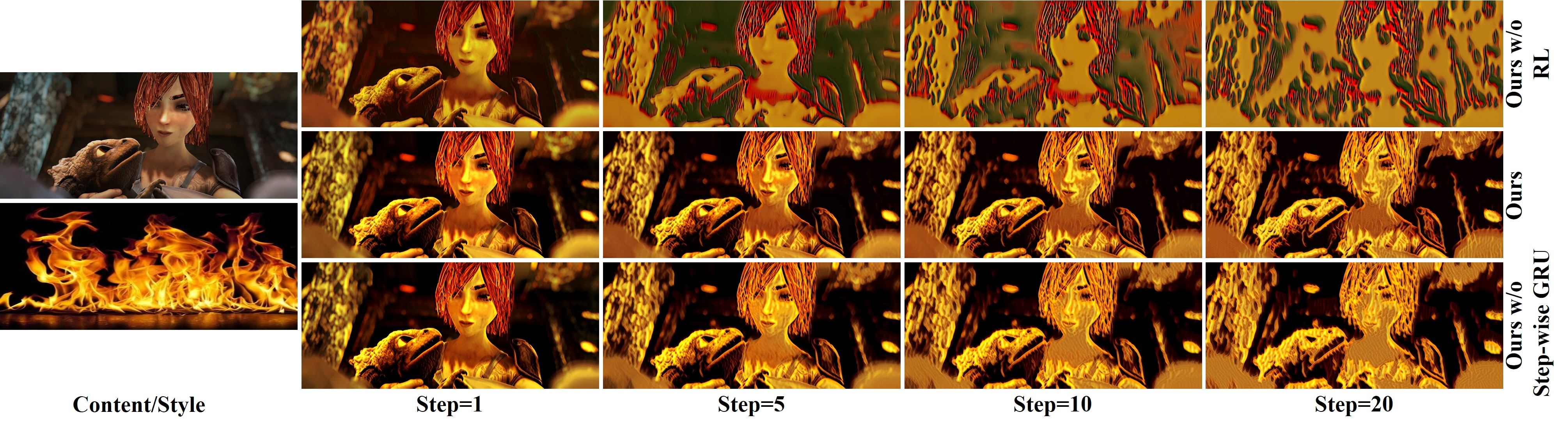}
    \caption{Comparison of our method, our method without using RL (AS method), and our method without using Step-wise GRU. RL makes the results from our model more stable and Step-wise GRU makes the output  has higher quality. }
    \label{fig:compare_ae_1_5}
\end{figure*}

\begin{table}[htb]
		\centering
            \small{
            \begin{tabular}{cc|c|c}
            \hline
            \multicolumn{2}{c|}{\diagbox{Methods}{Styles}}                                       & La\_muse          & Brushstrokes      \\ \hline
            \multicolumn{1}{c|}{\multirow{2}{*}{Step=1}}  & Ours w/o FWG & 1.8939          & 1.0933          \\
            \multicolumn{1}{c|}{}                         & Ours                    & \textbf{1.1351} & \textbf{0.8679} \\ \hline
            \multicolumn{1}{c|}{\multirow{2}{*}{Step=5}}  & Ours w/o FWG & 2.3991          & 1.4731          \\
            \multicolumn{1}{c|}{}                         & Ours                    & \textbf{1.8883} & \textbf{1.4331} \\ \hline
            \multicolumn{1}{c|}{\multirow{2}{*}{Step=10}} & Ours w/o FWG & 3.1329          & 1.8000          \\
            \multicolumn{1}{c|}{}                         & Ours                    & \textbf{2.3836} & \textbf{1.7053} \\ \hline
            \end{tabular}}
            \captionof{table}{Comparison of our method with and without using Frame-wise GRU (FWG). The average of temporal losses ($\times 10^{-2}$) from eight sequences are reported on two styles. 
            }
            \label{tab_abl_frame_gru}
\end{table}

\paragraph{Ablation Study.}
(1) We study the effect of the RL model in our framework. As shown in Fig.~\ref{compare_vae}, compared with the method that uses only Actor-Stylizer (AS), our method can generate more stable and clear stylized images at step 1. At step 10, AS loses the content information completely without the help of RL (Ours w/o RL), while our method can still produce amazing results. We also show the corresponding numerical comparison in Table~\ref{tab_abl}. We can easily see that our method achieves the best performance consistently in both steps 1 and 10.
This study indicates that RL can indeed improve the performance of DL-based NST models. (2) Since AdaIN and StyTR2 methods can adjust the hyperparameter `alpha'$\in[0,1]$ and the round of repetitive stylization to control the degree of stylization in the final results, respectively, we compare our method with them accordingly in Fig. \ref{compare_adain_stytr2}. From the visualization results, we can see that the results of AdaIN are in the under-stylized state even if the style control hyperparameter is changed. Moreover, StyTR2 gets the results with small style changes and low quality after multiple rounds. However, our method not only ensures the gradual change in style, but also produces very smooth results.

\subsection{Evaluations on Video RL-NST}

\paragraph{Qualitative Comparison.}
We show the visualization results of our method compared with the four latest video style transfer methods in Fig.~\ref{fig:compare_video}, wherein, for each method, the top portion shows the specific stylized results and the bottom portion is the heatmap of the differences in the adjacent frames of the input and stylized videos. Note that the adjacent frame indexes are the same for all methods. 
We can find that our method produces refined stylized results and our results are closest to the input frames. In particular, our method can highly promote the stability of video style transfer. The differences in our results are closest to the difference from input frames without reducing the effect of stylization. It is clear that MCCNet and ReReVST fail to keep the coherence of videos. In addition, Linear and AdaAttN also fail to keep the coherence in some regions that are close to the edge of objects such as the head and shoulder.

\paragraph{Quantitative Results.}
As shown in Table~\ref{tab:compare_video_methods}, we choose 23 different sequences from the MPI Sintel dataset~\cite{butler2012naturalistic} and eight different style images to calculate the average of temporal losses for comparison. 
It is clear that our method (step=1 and 5) outperforms the compared methods in all style settings. Our method still has a low temporal error even if step=10.

\paragraph{Ablation Study.}
We investigate the effect of the individual parts of the network structure group on the results, including the RL, Step-wise GRU, and Frame-wise GRU.

(1) As shown in Fig.~\ref{fig:compare_ae_1_5} (first and second rows), our method generates more stable and clearer stylized results than the method using only Actor-Stylizer without RL. After step 5, AS no longer has the ability to keep the content information and style information, while our method with RL can still produce good results. 
(2) Similarly, we have compared our method with the results produced when Step-wise GRU is removed, and the results are shown in Fig.~\ref{fig:compare_ae_1_5} (second and third rows). We can clearly see that most of the face details of the protagonist and dragon have been lost at step 10 when using our method without the Step-wise GRU. 
Also, the external details of the protagonist and dragon are completely lost in step 20. Our method with using Step-wise GRU, on the other hand, obtains very fine results even at step 20. 
(3) Table~\ref{tab_abl_frame_gru} shows the comparison of the temporal loss of our method with Frame-wise GRU (FWG) and without FWG. We find that the temporal loss is very low if we use FWG, which means the obtained final results are more consistent from frame to frame.
The above experiments have shown that RL, Step-wise , and Frame-wise GRU all greatly improve the performance of the model.

\section{Conclusion}
In this paper, we propose a new RL-based framework called RL-NST for both image and video NST tasks. 
It achieves style transfer step by step for a flexible control of the level of stylization. 
Despite using a lightweight neural network model, RL-NST can handle an extremely high-dimensional latent action space (up to 64$\times$64) and is thus capable of generating visually more satisfying artistic images than existing NST methods. 
Experimental results have demonstrated the effectiveness of the proposed method. 
The main goal of this work is to show the effectiveness of stylization-level controlling with our RL-based method and the superiority of our method in achieving the best NST quality. Therefore we use a single-style NST model. In the future, we would like to extend our model to multiple-style transfer tasks.

\section*{Acknowledgements}
This work was supported in part by Sichuan province Key Technology Research and Development project under Grant No. 2023YFG0305, 2023YFG0124, CUIT Science and Technology Innovation Capacity Enhancement Program Project under Grant KYTD202206.

\section*{Contribution Statement}
The contributions of Chengming Feng and Jing Hu to this paper were equal.


\bibliographystyle{named}
\bibliography{ijcai23}

\begin{thebibliography}{}

\bibitem[\protect\citeauthoryear{An \bgroup \em et al.\egroup
  }{2021}]{an2021artflow}
Jie An, Siyu Huang, Yibing Song, Dejing Dou, Wei Liu, and Jiebo Luo.
\newblock Artflow: Unbiased image style transfer via reversible neural flows.
\newblock In {\em CVPR}, pages 862--871, 2021.

\bibitem[\protect\citeauthoryear{Butler \bgroup \em et al.\egroup
  }{2012}]{butler2012naturalistic}
Daniel~J Butler, Jonas Wulff, Garrett~B Stanley, and Michael~J Black.
\newblock A naturalistic open source movie for optical flow evaluation.
\newblock In {\em ECCV}, pages 611--625. Springer, 2012.

\bibitem[\protect\citeauthoryear{Chen \bgroup \em et al.\egroup
  }{2017}]{chen2017stylebank}
Dongdong Chen, Lu~Yuan, Jing Liao, Nenghai Yu, and Gang Hua.
\newblock Stylebank: An explicit representation for neural image style
  transfer.
\newblock In {\em CVPR}, 2017.

\bibitem[\protect\citeauthoryear{Chen \bgroup \em et al.\egroup
  }{2021}]{chen2021artistic}
Haibo Chen, Zhizhong Wang, Huiming Zhang, Zhiwen Zuo, Ailin Li, Wei Xing,
  Dongming Lu, et~al.
\newblock Artistic style transfer with internal-external learning and
  contrastive learning.
\newblock {\em Advances in Neural Information Processing Systems}, 34, 2021.

\bibitem[\protect\citeauthoryear{Cheng \bgroup \em et al.\egroup
  }{2021}]{cheng2021style}
Jiaxin Cheng, Ayush Jaiswal, Yue Wu, Pradeep Natarajan, and Prem Natarajan.
\newblock Style-aware normalized loss for improving arbitrary style transfer.
\newblock In {\em CVPR}, 2021.

\bibitem[\protect\citeauthoryear{Deng \bgroup \em et al.\egroup
  }{2021a}]{deng2020arbitrary}
Yingying Deng, Fan Tang, Weiming Dong, Haibin Huang, Chongyang Ma, and
  Changsheng Xu.
\newblock Arbitrary video style transfer via multi-channel correlation.
\newblock {\em AAAI}, 2021.

\bibitem[\protect\citeauthoryear{Deng \bgroup \em et al.\egroup
  }{2021b}]{deng2021stytr}
Yingying Deng, Fan Tang, Xingjia Pan, Weiming Dong, Chongyang Ma, and
  Changsheng Xu.
\newblock Stytr\^{} 2: Unbiased image style transfer with transformers.
\newblock {\em arXiv preprint arXiv:2105.14576}, 2021.

\bibitem[\protect\citeauthoryear{Donahue \bgroup \em et al.\egroup
  }{2017}]{7558228}
Jeff Donahue, Lisa~Anne Hendricks, Marcus Rohrbach, Subhashini Venugopalan,
  Sergio Guadarrama, Kate Saenko, and Trevor Darrell.
\newblock Long-term recurrent convolutional networks for visual recognition and
  description.
\newblock {\em IEEE Transactions on Pattern Analysis and Machine Intelligence},
  39(4):677--691, 2017.

\bibitem[\protect\citeauthoryear{Gatys \bgroup \em et al.\egroup
  }{2015a}]{gatys2015texture}
Leon Gatys, Alexander~S Ecker, and Matthias Bethge.
\newblock Texture synthesis using convolutional neural networks.
\newblock {\em NeurIPS}, 28:262--270, 2015.

\bibitem[\protect\citeauthoryear{Gatys \bgroup \em et al.\egroup
  }{2015b}]{gatys2015neural}
Leon~A Gatys, Alexander~S Ecker, and Matthias Bethge.
\newblock A neural algorithm of artistic style.
\newblock {\em arXiv preprint arXiv:1508.06576}, 2015.

\bibitem[\protect\citeauthoryear{Gu \bgroup \em et al.\egroup
  }{2017}]{gu2017deep}
Shixiang Gu, Ethan Holly, Timothy Lillicrap, and Sergey Levine.
\newblock Deep reinforcement learning for robotic manipulation with
  asynchronous off-policy updates.
\newblock In {\em 2017 IEEE ICRA}, pages 3389--3396. IEEE, 2017.

\bibitem[\protect\citeauthoryear{Haarnoja \bgroup \em et al.\egroup
  }{2017}]{haarnoja2017reinforcement}
Tuomas Haarnoja, Haoran Tang, Pieter Abbeel, and Sergey Levine.
\newblock Reinforcement learning with deep energy-based policies.
\newblock In {\em ICML}, pages 1352--1361. PMLR, 2017.

\bibitem[\protect\citeauthoryear{Haarnoja \bgroup \em et al.\egroup
  }{2018}]{haarnoja2018soft}
Tuomas Haarnoja, Aurick Zhou, Kristian Hartikainen, George Tucker, Sehoon Ha,
  Jie Tan, Vikash Kumar, Henry Zhu, Abhishek Gupta, Pieter Abbeel, et~al.
\newblock Soft actor-critic algorithms and applications.
\newblock {\em arXiv preprint arXiv:1812.05905}, 2018.

\bibitem[\protect\citeauthoryear{He \bgroup \em et al.\egroup
  }{2016}]{he2016deep}
Kaiming He, Xiangyu Zhang, Shaoqing Ren, and Jian Sun.
\newblock Deep residual learning for image recognition.
\newblock In {\em CVPR}, pages 770--778, 2016.

\bibitem[\protect\citeauthoryear{Hu \bgroup \em et al.\egroup
  }{2023}]{hu2023attention}
Jing Hu, Zhikun Shuai, Xin Wang, Shu Hu, Shanhui Sun, Siwei Lyu, and Xi~Wu.
\newblock Attention guided policy optimization for 3d medical image
  registration.
\newblock {\em IEEE Access}, 2023.

\bibitem[\protect\citeauthoryear{Huang and Belongie}{2017}]{huang2017arbitrary}
Xun Huang and Serge Belongie.
\newblock Arbitrary style transfer in real-time with adaptive instance
  normalization.
\newblock In {\em Proceedings of the IEEE ICCV}, pages 1501--1510, 2017.

\bibitem[\protect\citeauthoryear{Johnson \bgroup \em et al.\egroup
  }{2016}]{johnson2016perceptual}
Justin Johnson, Alexandre Alahi, and Li~Fei-Fei.
\newblock Perceptual losses for real-time style transfer and super-resolution.
\newblock In {\em ECCV}. Springer, 2016.

\bibitem[\protect\citeauthoryear{Kingma and Welling}{2013}]{kingma2013auto}
Diederik~P Kingma and Max Welling.
\newblock Auto-encoding variational bayes.
\newblock {\em arXiv preprint arXiv:1312.6114}, 2013.

\bibitem[\protect\citeauthoryear{Li \bgroup \em et al.\egroup
  }{2017a}]{li2017demystifying}
Yanghao Li, Naiyan Wang, Jiaying Liu, and Xiaodi Hou.
\newblock Demystifying neural style transfer.
\newblock {\em arXiv preprint arXiv:1701.01036}, 2017.

\bibitem[\protect\citeauthoryear{Li \bgroup \em et al.\egroup
  }{2017b}]{li2017universal}
Yijun Li, Chen Fang, Jimei Yang, Zhaowen Wang, Xin Lu, and Ming-Hsuan Yang.
\newblock Universal style transfer via feature transforms.
\newblock {\em Advances in neural information processing systems}, 30, 2017.

\bibitem[\protect\citeauthoryear{Li \bgroup \em et al.\egroup
  }{2019}]{li2019learning}
Xueting Li, Sifei Liu, Jan Kautz, and Ming-Hsuan Yang.
\newblock Learning linear transformations for fast image and video style
  transfer.
\newblock In {\em Proceedings of the IEEE/CVF Conference on Computer Vision and
  Pattern Recognition}, pages 3809--3817, 2019.

\bibitem[\protect\citeauthoryear{Lillicrap \bgroup \em et al.\egroup
  }{2015}]{lillicrap2015continuous}
Timothy~P Lillicrap, Jonathan~J Hunt, Alexander Pritzel, Nicolas Heess, Tom
  Erez, Yuval Tassa, David Silver, and Daan Wierstra.
\newblock Continuous control with deep reinforcement learning.
\newblock {\em arXiv preprint arXiv:1509.02971}, 2015.

\bibitem[\protect\citeauthoryear{Lin \bgroup \em et al.\egroup
  }{2014}]{lin2014microsoft}
Tsung-Yi Lin, Michael Maire, Serge Belongie, James Hays, Pietro Perona, Deva
  Ramanan, Piotr Doll{\'a}r, and C~Lawrence Zitnick.
\newblock Microsoft coco: Common objects in context.
\newblock In {\em ECCV}, pages 740--755. Springer, 2014.

\bibitem[\protect\citeauthoryear{Lin \bgroup \em et al.\egroup
  }{2021}]{lin2021drafting}
Tianwei Lin, Zhuoqi Ma, Fu~Li, Dongliang He, Xin Li, Errui Ding, Nannan Wang,
  Jie Li, and Xinbo Gao.
\newblock Drafting and revision: Laplacian pyramid network for fast
  high-quality artistic style transfer.
\newblock In {\em CVPR}, pages 5141--5150, 2021.

\bibitem[\protect\citeauthoryear{Liu \bgroup \em et al.\egroup
  }{2021}]{liu2021adaattn}
Songhua Liu, Tianwei Lin, Dongliang He, Fu~Li, Meiling Wang, Xin Li, Zhengxing
  Sun, Qian Li, and Errui Ding.
\newblock Adaattn: Revisit attention mechanism in arbitrary neural style
  transfer.
\newblock In {\em Proceedings of the IEEE/CVF International Conference on
  Computer Vision}, pages 6649--6658, 2021.

\bibitem[\protect\citeauthoryear{Luo \bgroup \em et al.\egroup
  }{2021}]{luostochastic}
Ziwei Luo, Jing Hu, Xin Wang, Siwei Lyu, Bin Kong, Youbing Yin, Qi~Song, and
  Xi~Wu.
\newblock Stochastic actor-executor-critic for image-to-image translation.
\newblock {\em IJCAI}, 2021.

\bibitem[\protect\citeauthoryear{Mahendran and
  Vedaldi}{2015}]{mahendran2015understanding}
Aravindh Mahendran and Andrea Vedaldi.
\newblock Understanding deep image representations by inverting them.
\newblock In {\em CVPR}, pages 5188--5196, 2015.

\bibitem[\protect\citeauthoryear{Mirowski \bgroup \em et al.\egroup
  }{2017}]{mirowski2016learning}
Piotr Mirowski, Razvan Pascanu, Fabio Viola, Hubert Soyer, Andrew~J Ballard,
  Andrea Banino, Misha Denil, Ross Goroshin, Laurent Sifre, Koray Kavukcuoglu,
  et~al.
\newblock Learning to navigate in complex environments.
\newblock {\em ICLR}, 2017.

\bibitem[\protect\citeauthoryear{Park and Lee}{2019}]{park2019arbitrary}
Dae~Young Park and Kwang~Hee Lee.
\newblock Arbitrary style transfer with style-attentional networks.
\newblock In {\em proceedings of the IEEE/CVF conference on computer vision and
  pattern recognition}, pages 5880--5888, 2019.

\bibitem[\protect\citeauthoryear{pex}{2022}]{pexels2022}
Pexels.
\newblock \url{https://www.pexels.com/}, 2022.
\newblock Accessed: 2022-03-12.

\bibitem[\protect\citeauthoryear{Phillips and
  Mackintosh}{2011}]{phillips2011wiki}
Fred Phillips and Brandy Mackintosh.
\newblock Wiki art gallery, inc.: A case for critical thinking.
\newblock {\em Issues in Accounting Education}, 26(3):593--608, 2011.

\bibitem[\protect\citeauthoryear{Risser \bgroup \em et al.\egroup
  }{2017}]{risser2017stable}
Eric Risser, Pierre Wilmot, and Connelly Barnes.
\newblock Stable and controllable neural texture synthesis and style transfer
  using histogram losses.
\newblock {\em arXiv preprint arXiv:1701.08893}, 2017.

\bibitem[\protect\citeauthoryear{Sheng \bgroup \em et al.\egroup
  }{2018}]{sheng2018avatar}
Lu~Sheng, Ziyi Lin, Jing Shao, and Xiaogang Wang.
\newblock Avatar-net: Multi-scale zero-shot style transfer by feature
  decoration.
\newblock In {\em Proceedings of the IEEE conference on computer vision and
  pattern recognition}, pages 8242--8250, 2018.

\bibitem[\protect\citeauthoryear{Shi \bgroup \em et al.\egroup
  }{2015}]{shi2015convolutional}
Xingjian Shi, Zhourong Chen, Hao Wang, Dit-Yan Yeung, Wai-Kin Wong, and
  Wang-chun Woo.
\newblock Convolutional lstm network: A machine learning approach for
  precipitation nowcasting.
\newblock {\em Advances in neural information processing systems}, 28, 2015.

\bibitem[\protect\citeauthoryear{Simonyan and
  Zisserman}{2014}]{simonyan2014very}
Karen Simonyan and Andrew Zisserman.
\newblock Very deep convolutional networks for large-scale image recognition.
\newblock {\em arXiv preprint arXiv:1409.1556}, 2014.

\bibitem[\protect\citeauthoryear{Tassa \bgroup \em et al.\egroup
  }{2018}]{tassa2018deepmind}
Yuval Tassa, Yotam Doron, Alistair Muldal, Tom Erez, Yazhe Li, Diego de~Las
  Casas, David Budden, Abbas Abdolmaleki, Josh Merel, Andrew Lefrancq, et~al.
\newblock Deepmind control suite.
\newblock {\em arXiv preprint arXiv:1801.00690}, 2018.

\bibitem[\protect\citeauthoryear{Ulyanov \bgroup \em et al.\egroup
  }{2016a}]{ulyanov2016texture}
Dmitry Ulyanov, Vadim Lebedev, Andrea Vedaldi, and Victor~S Lempitsky.
\newblock Texture networks: Feed-forward synthesis of textures and stylized
  images.
\newblock In {\em ICML}, volume~1, page~4, 2016.

\bibitem[\protect\citeauthoryear{Ulyanov \bgroup \em et al.\egroup
  }{2016b}]{ulyanov2016instance}
Dmitry Ulyanov, Andrea Vedaldi, and Victor Lempitsky.
\newblock Instance normalization: The missing ingredient for fast stylization.
\newblock {\em arXiv preprint arXiv:1607.08022}, 2016.

\bibitem[\protect\citeauthoryear{Wang \bgroup \em et al.\egroup
  }{2020}]{wang2020consistent}
Wenjing Wang, Shuai Yang, Jizheng Xu, and Jiaying Liu.
\newblock Consistent video style transfer via relaxation and regularization.
\newblock {\em IEEE Transactions on Image Processing}, 29:9125--9139, 2020.

\bibitem[\protect\citeauthoryear{Xiang \bgroup \em et al.\egroup
  }{2022}]{xiang2022rmbench}
Yanfei Xiang, Xin Wang, Shu Hu, Bin Zhu, Xiaomeng Huang, Xi~Wu, and Siwei Lyu.
\newblock Rmbench: Benchmarking deep reinforcement learning for robotic
  manipulator control.
\newblock {\em arXiv preprint arXiv:2210.11262}, 2022.

\bibitem[\protect\citeauthoryear{Zhao \bgroup \em et al.\egroup
  }{2019}]{zhao2019uncertainty}
Xujiang Zhao, Shu Hu, Jin-Hee Cho, and Feng Chen.
\newblock Uncertainty-based decision making using deep reinforcement learning.
\newblock In {\em 2019 22th International Conference on Information Fusion
  (FUSION)}, pages 1--8. IEEE, 2019.

\end{thebibliography}

\end{document}